\documentclass[10pt,twocolumn,letterpaper]{article}

\usepackage{cvpr}
\usepackage{times}
\usepackage{epsfig}
\usepackage{graphicx}
\usepackage{amsmath}
\usepackage{amssymb}
\usepackage{subfig}
\usepackage{caption}

\usepackage[breaklinks=true,bookmarks=false]{hyperref}

\cvprfinalcopy 

\def\cvprPaperID{****} 

\ifcvprfinal\fi
\begin{document}

\title{Extending Adversarial Attacks and Defenses to Deep 3D Point Cloud Classifiers}

\author{Daniel Liu\\
Torrey Pines High School\\
San Diego, CA\\
{\tt\small daniel.liu02@gmail.com}
\and
Ronald Yu\\
UCSD\\
La Jolla, CA\\
{\tt\small ronaldyu@ucsd.edu}
\and
Hao Su\\
UCSD\\
La Jolla, CA\\
{\tt\small haosu@eng.ucsd.edu}
}

\maketitle

\begin{abstract}
    3D object classification and segmentation using deep neural networks has been extremely successful. As the problem of identifying 3D objects has many safety-critical applications, the neural networks have to be robust against adversarial changes to the input data set. There is a growing body of research on generating human-imperceptible adversarial attacks and defenses against them in the 2D image classification domain. However, 3D objects have various differences with 2D images, and this specific domain has not been rigorously studied so far.
    
    We present a preliminary evaluation of adversarial attacks on deep 3D point cloud classifiers, namely PointNet and PointNet++, by evaluating both white-box and black-box adversarial attacks that were proposed for 2D images and extending those attacks to reduce the perceptibility of the perturbations in 3D space. We also show the high effectiveness of simple defenses against those attacks by proposing new defenses that exploit the unique structure of 3D point clouds. Finally, we attempt to explain the effectiveness of the defenses through the intrinsic structures of both the point clouds and the neural network architectures. Overall, we find that networks that process 3D point cloud data are weak to adversarial attacks, but they are also more easily defensible compared to 2D image classifiers. Our investigation will provide the groundwork for future studies on improving the robustness of deep neural networks that handle 3D data.
\end{abstract}

\section{Introduction}

Recent advances in 3D deep learning have made strides in tasks previously established by 2D baselines such as classification~\cite{qi2017pointnet}, segmentation~\cite{wang2018sgpn}, and detection~\cite{qi2017frustum}. However, 3D deep learning literature also lags behind its 2D counterpart on tasks that seek to better understand behavior of deep neural networks such as network interpretation~\cite{zhou2017interpreting}, few-shot learning~\cite{lake2015human}, and robustness to adversarial examples~\cite{goodfellow2014explaining}. 
We provide a preliminary investigation into how deep 3D neural networks behave by examining its behavior on simple adversarial attacks that are extremely effective on 2D images. We focus on examining networks, like PointNet~\cite{qi2017pointnet} and PointNet++~\cite{qi2017pointnetplusplus}, that process the lightweight point cloud representation of 3D objects.

Robustness against adversarial attacks has been subject to rigorous research due to its security implications in deep learning systems. Deep neural networks that process 2D images were shown to be extremely vulnerable against simple adversarial perturbations~\cite{goodfellow2014explaining}. Furthermore, many proposed defense methods have been foiled by adversarial attacks, which indicates the difficulty of the challenge posed by adversarial attacks~\cite{carlini2017adversarial}. These attacks are \textit{imperceptible} to humans, yet extremely effective in fooling neural networks. Attacks were also shown to be effectively transferable across different neural networks~\cite{kurakin2016adversarialscale} in black-box attacks (as opposed to white-box attacks, where the adversary has the model and its trained parameters).

We seek to advance studies in both 3D shape classification and adversarial robustness by examining the behavior of deep learning on point clouds in an adversarial setting. We evaluate the PointNet~\cite{qi2017pointnet} and PointNet++~\cite{qi2017pointnetplusplus} frameworks that apply shared multi-layer perceptrons on each point before using a global max-pooling layer across all the points to obtain a global feature vector of the entire shape.

In this paper, we achieve the following:

\begin{itemize}
    \item We show that various white-box and black-box adversarial attacks are effective on undefended point cloud classifiers.
    \item We show that simple defenses are effective against the white-box adversarial attacks.
    \item We discuss potential reasons behind the effectiveness of the defenses, including both intrinsic properties of the point clouds and the neural networks.
\end{itemize}
We adapt adversarial attacks and defenses for 3D point clouds, and we find that deep 3D point cloud classifiers, while susceptible to simple adversarial attacks, are also more easily defended than its 2D counterparts.

\section{Related works}
\subsection{Adversarial examples}
There have been a lot of research on both generating and defending against adversarial attacks. Attacks on convolutional neural networks for 2D image classification is the most popular, with~\cite{szegedy2013intriguing} first introducing an optimization-based attack, and Goodfellow \etal~\cite{goodfellow2014explaining} proposing a simple and efficient fast gradient sign method (FGSM) for generating adversarial attacks constrained by the $L_\infty$ norm. Extensions to the fast gradient sign method include running it for multiple iterations \cite{kurakin2016adversarialphysical} and using momentum \cite{dong2018boosting}. Other effective attacks include the Jacobian-based saliency map attack (JSMA)~\cite{papernot2016limitations}, DeepFool~\cite{moosavi2016deepfool}, and the Carlini-Wagner attack~\cite{carlini2017towards}.

The effectiveness of adversarial attacks has also been examined in other domains. \cite{biggio2013evasion} proposed an optimization-based attack against malware detection. \cite{arnab2017robustness} evaluated adversarial attacks on 2D image segmentation. Only very recently has there been work on examining adversarial point clouds~\cite{xiang2018generating, zhou2018deflecting, zheng2018learning}.

There has been many techniques proposed for defense, including adversarial training~\cite{goodfellow2014explaining} and defensive distillation~\cite{papernot2016distillation}.

\subsection{3D deep learning}
3D shape classification has been studied for various representations of 3D objects: point clouds~\cite{deng2018ppf, qi2017pointnetplusplus, qi2017pointnet}, meshes~\cite{yi2017syncspeccnn}, and voxels~\cite{wang2017cnn}.
\section{White-box adversarial attacks}
\begin{figure*}[!htbp]
    \centering
    \captionsetup[subfigure]{width=0.2\textwidth}
    \subfloat[Original car]{\includegraphics[trim={8cm 3cm 8cm 3cm}, clip=true, width=0.25\linewidth]{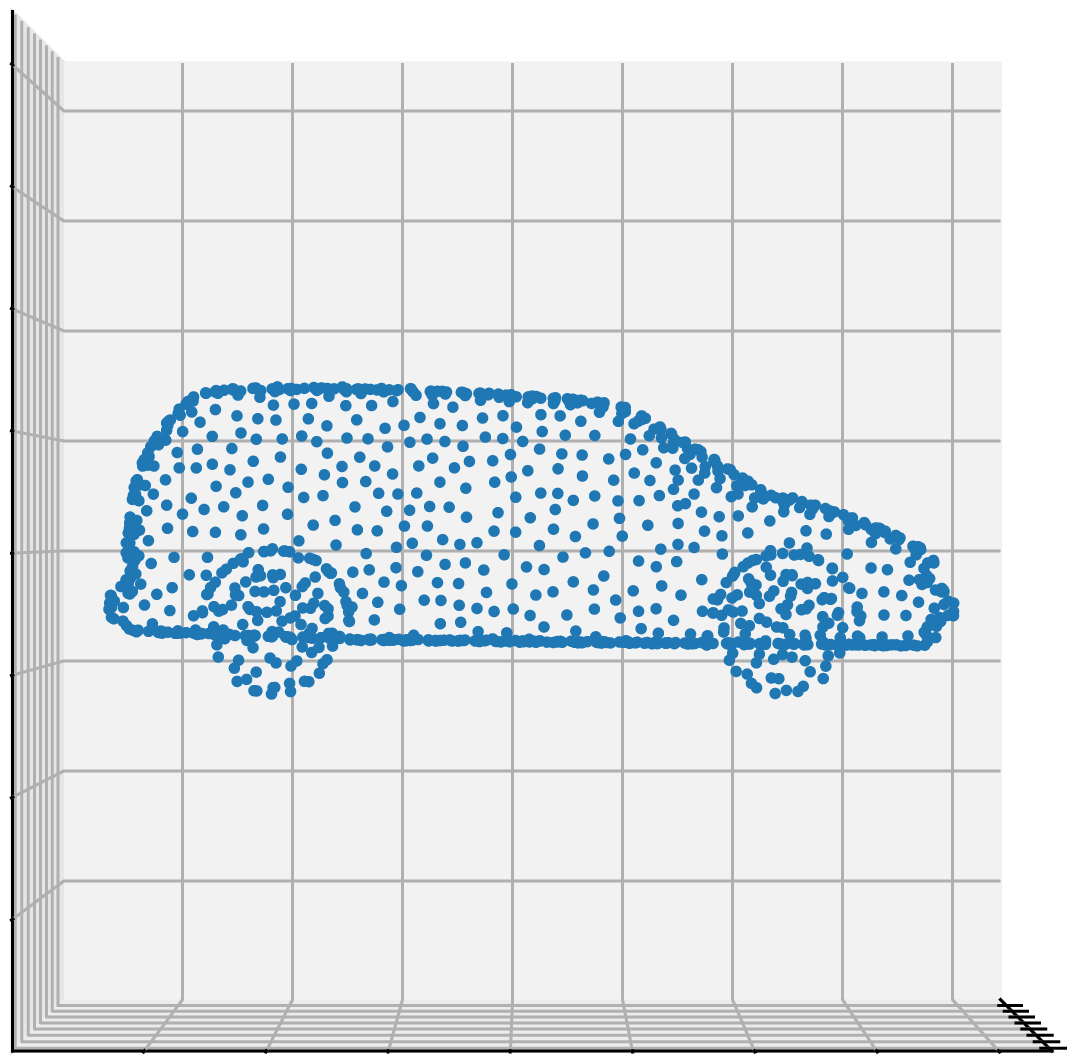}}
    \subfloat[Fast gradient $L_2$, predicted as bookshelf.]{\includegraphics[trim={8cm 3cm 8cm 3cm}, clip=true, width=0.25\linewidth]{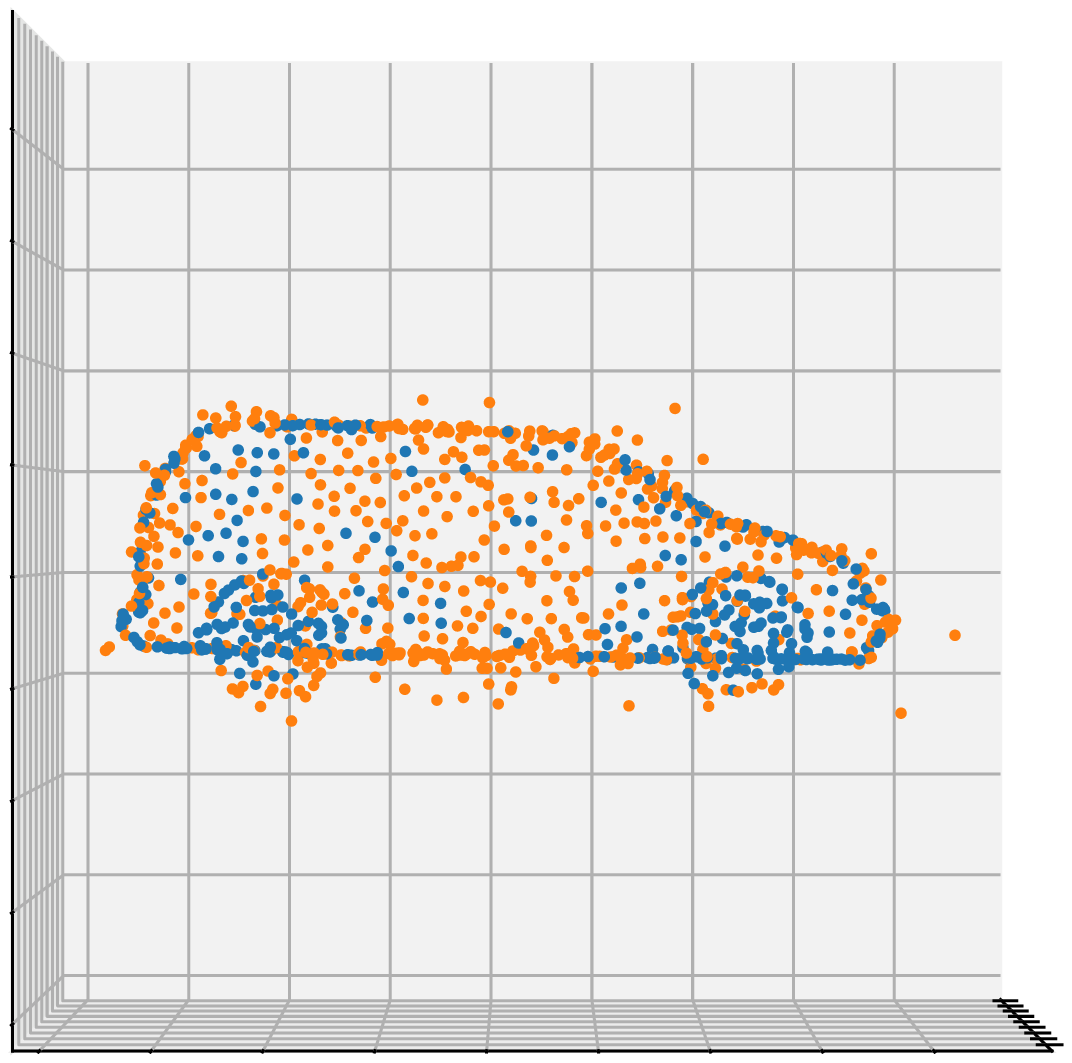}}
    \subfloat[Iter. gradient $L_2$, predicted as range hood.]{\includegraphics[trim={8cm 3cm 8cm 3cm}, clip=true, width=0.25\linewidth]{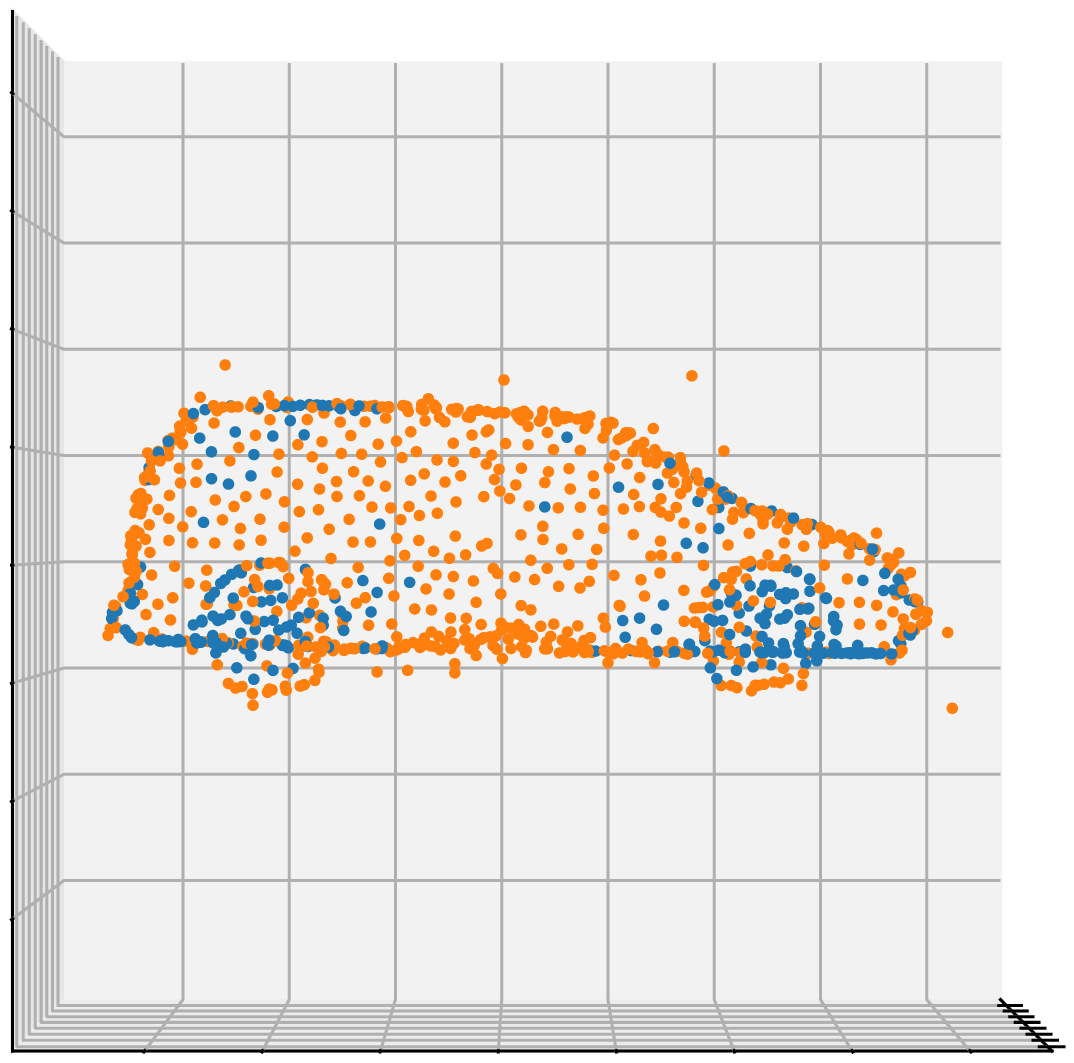}}
    \subfloat[Iter. gradient $L_2$ and clipping norms, predicted as range hood.]{\includegraphics[trim={8cm 3cm 8cm 3cm}, clip=true, width=0.25\linewidth]{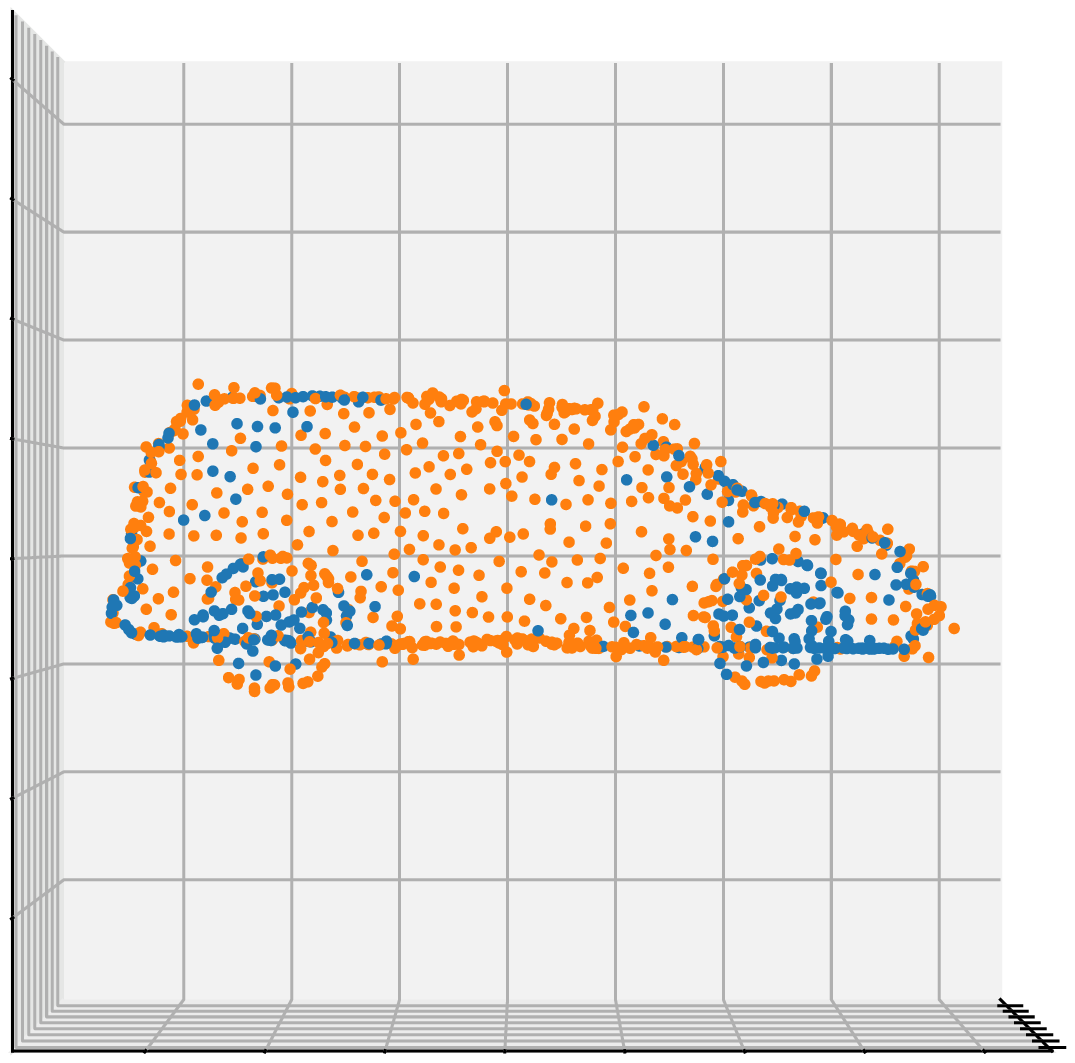}}
    \vfill
    \subfloat[Iter. gradient $L_2$ and gradient proj., predicted as range hood.]{\includegraphics[trim={8cm 3cm 8cm 3cm}, clip=true, width=0.25\linewidth]{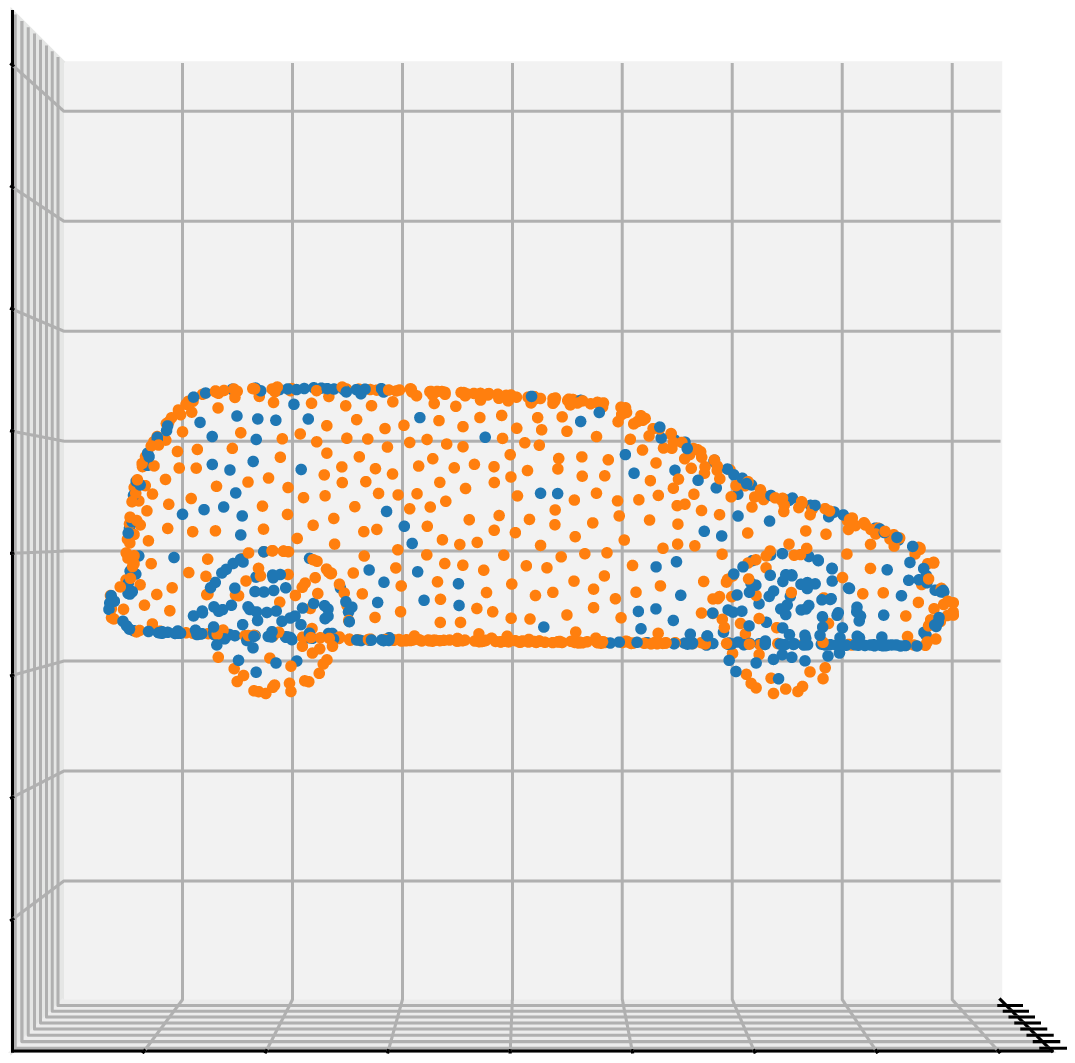}}
    \subfloat[Normalized fast gradient $L_2$, predicted as bookshelf.]{\includegraphics[trim={8cm 3cm 8cm 3cm}, clip=true, width=0.25\linewidth]{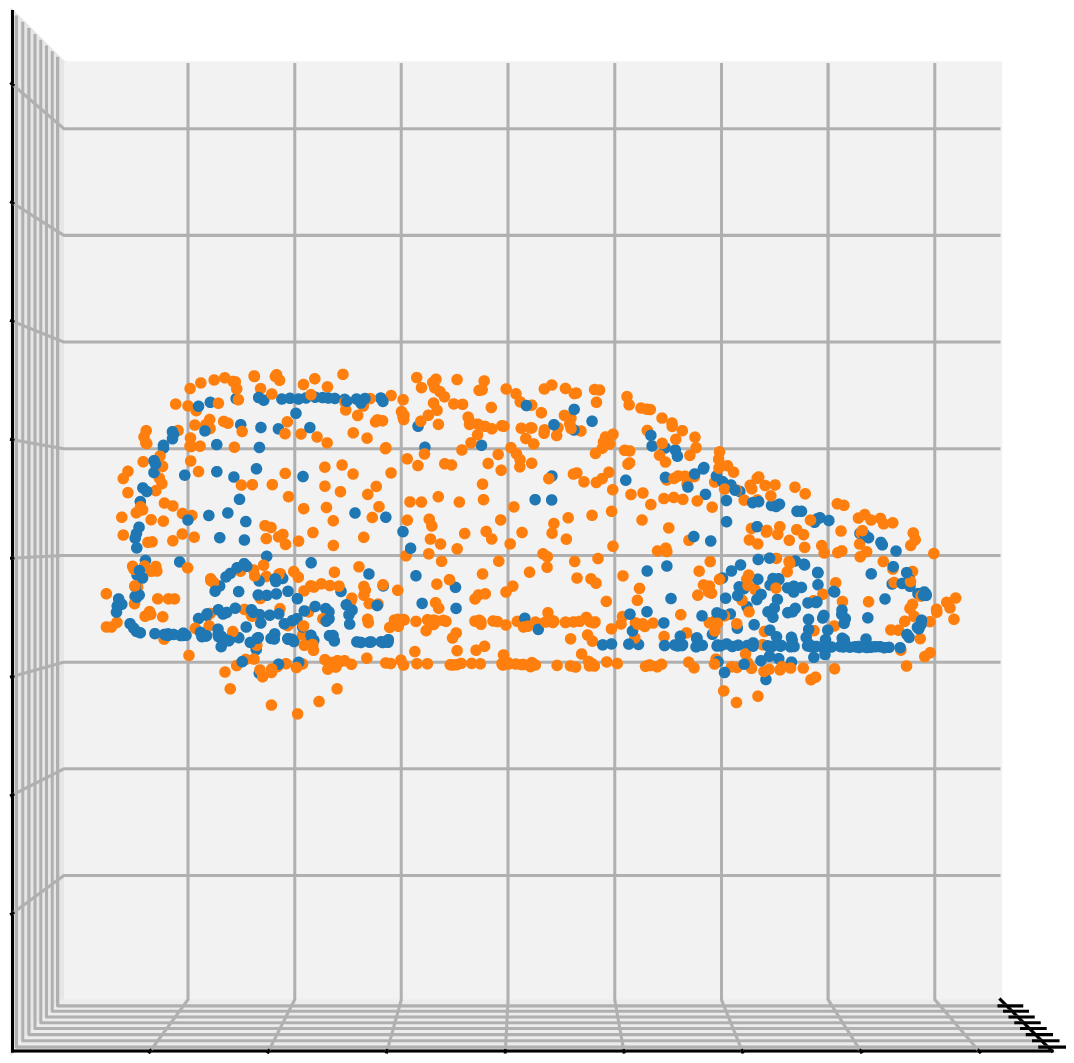}}
    \subfloat[Normalized iter. gradient $L_2$, predicted as range hood.]{\includegraphics[trim={8cm 3cm 8cm 3cm}, clip=true, width=0.25\linewidth]{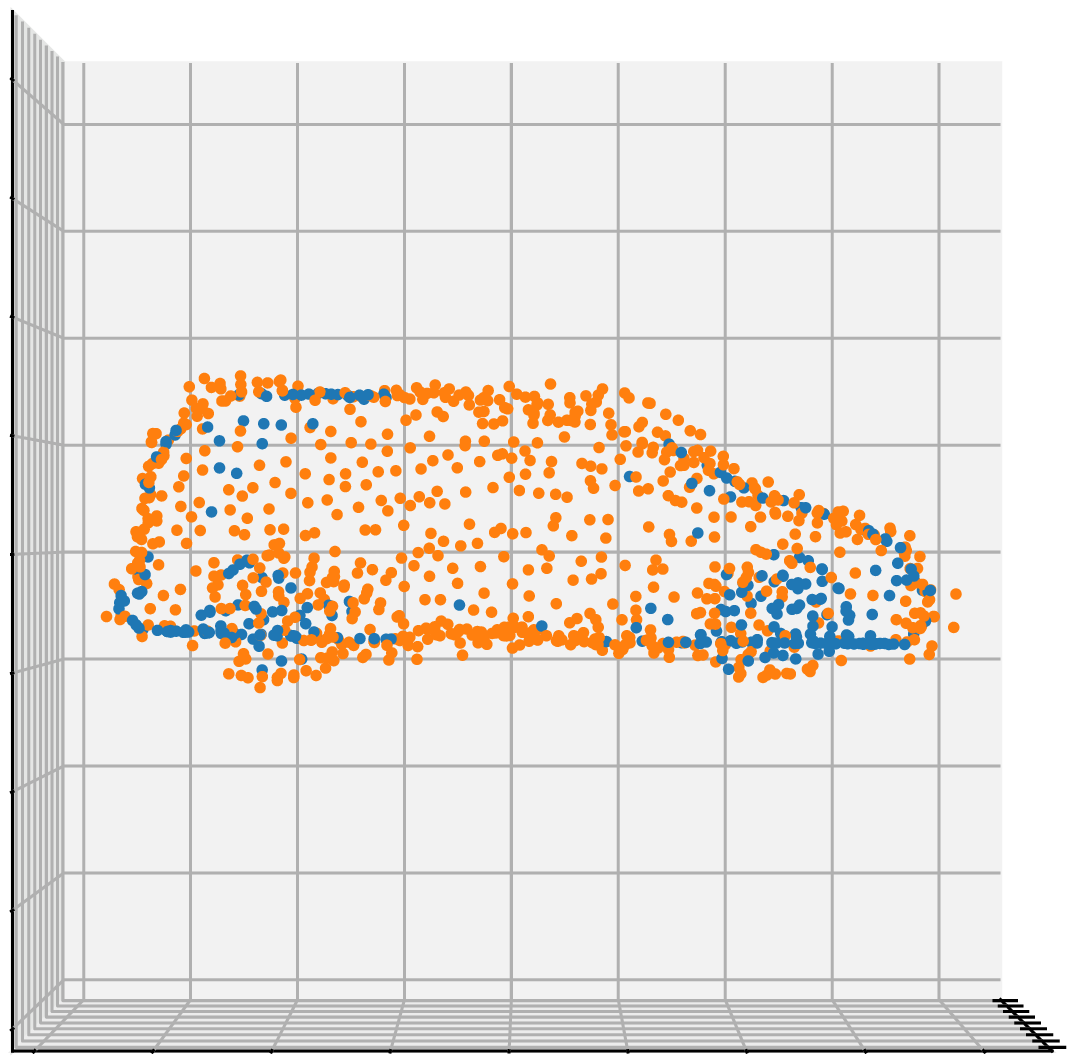}}
    \caption{A set of successful adversarial perturbations on the point cloud of a car, generated from PointNet trained with the ModelNet-Unique dataset. Orange points have nonzero perturbations.}
    \label{fig:adv_atk_viz_car}
\end{figure*}

As a preliminary investigation on the robustness of 3D deep neural networks to adversarial examples, we explore both targeted (\ie, misclassify to a specific class) and untargeted (\ie, misclassify to any class other than correct label) adversarial attacks.

\subsection{Fast gradient method}
The fast gradient sign method (FGSM) introduced by Goodfellow \etal~\cite{goodfellow2014explaining} generates adversarial examples for a deep neural network $f$ (that is parameterized by $\theta$ and takes an input $x$) by increasing its cross entropy loss $J$ between the network's output and the label $y$ while constraining the $L_\infty$ norm of the perturbation of $x$:
\begin{equation}
x^{adv} = x + \epsilon \operatorname{sign}(\Delta_x J(f(x; \theta), y))
\end{equation}
The $\epsilon$ value is an adjustable hyperparameter that dictates the $L_\infty$ norm between the original input and the adversarial example (\ie, $||x^{adv} - x||_\infty \leq \epsilon$). For targeted attacks, the sign of the gradient is subtracted instead of added to the original sample.

\subsection{Iterative gradient method}
The iterative fast gradient method~\cite{kurakin2016adversarialphysical} improves the fast gradient attack by repeating it multiple times to get a better estimate of the loss surface wrt to the input of the network. The algorithm for crafting an adversarial example constrained by the $L_\infty$ norm can be expressed as
\begin{align}
\begin{split}
x_0^{adv} &= x\\
x_t^{adv} &= x_{t - 1}^{adv} + \epsilon \operatorname{sign}(\Delta_{x_{t - 1}^{adv}} J(f(x_{t - 1}^{adv}; \theta), y))
\end{split}
\end{align}

\subsection{Modifying the fast/iterative gradient method}
We expand Goodfellow \etal's~\cite{goodfellow2014explaining} idea to several related categories of attacks. All of these cases constrain the magnitude of the perturbation onto the surface of an epsilon ball, but in different dimensions.

\begin{itemize}
    \item Constraining the $L_2$ norm of the perturbation for each dimension of each point. This is just Goodfellow \etal's FGSM~\cite{goodfellow2014explaining}. It restricts each dimension's perturbation onto the surface of an 1D epsilon ball by using the sign operation on all the points.
    \item Constraining the $L_2$ norm of the perturbation for each point. We do this by normalizing all 3 dimensions of each point's perturbation by its $L_2$ norm. This constrains the perturbation for each point onto the surface of a 3D epsilon ball, allowing it to be in any arbitrary direction. We refer to this as the "normalized gradient $L_2$ method". Each point $p \in x$ is perturbed with the following equation:
    \begin{equation}
    p^{adv} = p + \epsilon \frac{\Delta_p J(f(x; \theta), y)}{||\Delta_p J(f(x; \theta), y)||_2}
    \end{equation}
    \item Constraining the $L_2$ norm between the entire clean point cloud and the entire adversarial point cloud. This was explored for 2D images by \cite{kurakin2016adversarialscale} and \cite{miyato2015distributional}. We do this by normalizing each dimension's perturbation by the $L_2$ norm of the perturbation for all dimensions. This allows the individual perturbations to have diverse magnitudes and directions. We refer to this as the "gradient $L_2$ method", and it is formally defined as
    \begin{equation}
    x^{adv} = x + \epsilon \frac{\Delta_x J(f(x; \theta), y)}{||\Delta_x J(f(x; \theta), y)||_2}
    \end{equation}
\end{itemize}

Our preliminary tests have shown little difference between the iterative attack success rates of all three methods. However, in terms of perceptibility, each attack variation affects the point cloud differently. Constraining the $L_\infty$ norm of the perturbation severely limits the number of perturbation directions in 3D space due to the sign operation. By allowing various perturbation magnitudes and directions with the gradient $L_2$ method, the attack can create outliers by assigning higher perturbations to input features with high gradients and allow the points that have lower gradients to be perturbed less. Normalized gradient $L_2$ allows more available perturbation directions for each point, but it does not generate outliers.
We will mainly consider the latter two variations in our evaluations, as the first method is much more restricted regarding how each point can be perturbed.

\begin{figure*}[!htbp]
    \centering
    \captionsetup[subfigure]{width=0.2\textwidth}
    \subfloat[Original person]{\includegraphics[trim={8cm 3cm 8cm 3cm}, clip=true, width=0.25\linewidth]{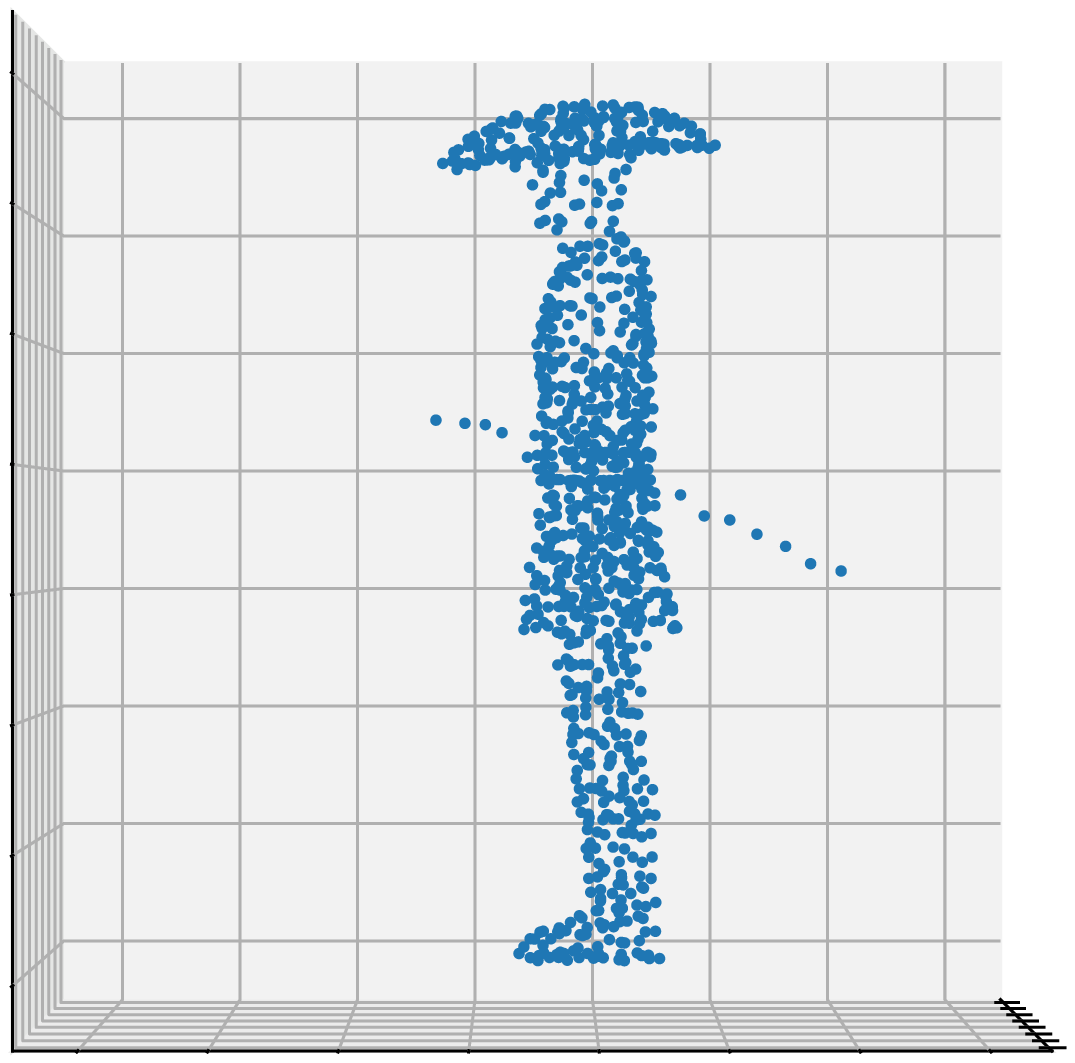}}
    \subfloat[Fast gradient $L_2$, predicted as plant.]{\includegraphics[trim={8cm 3cm 8cm 3cm}, clip=true, width=0.25\linewidth]{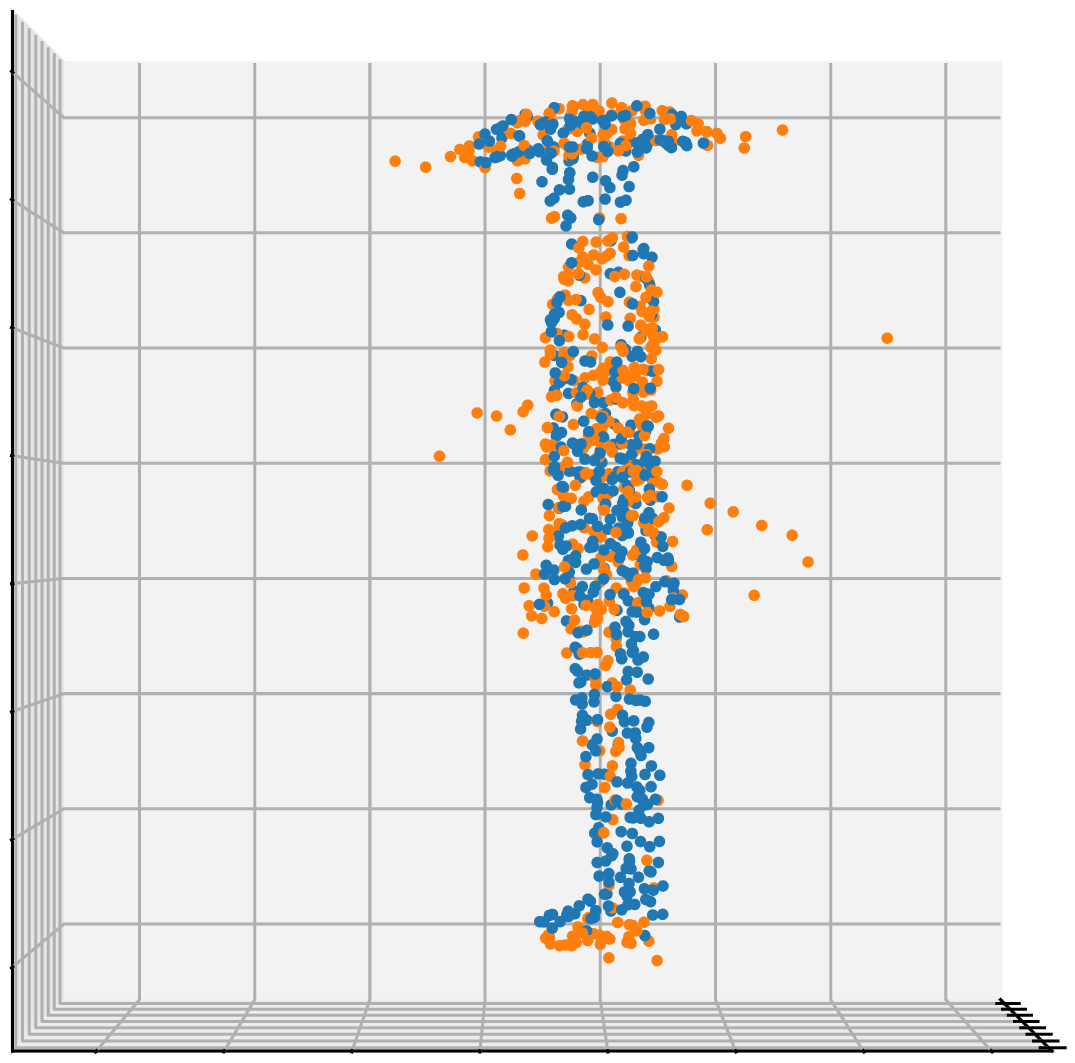}}
    \subfloat[Iter. gradient $L_2$, predicted as plant.]{\includegraphics[trim={8cm 3cm 8cm 3cm}, clip=true, width=0.25\linewidth]{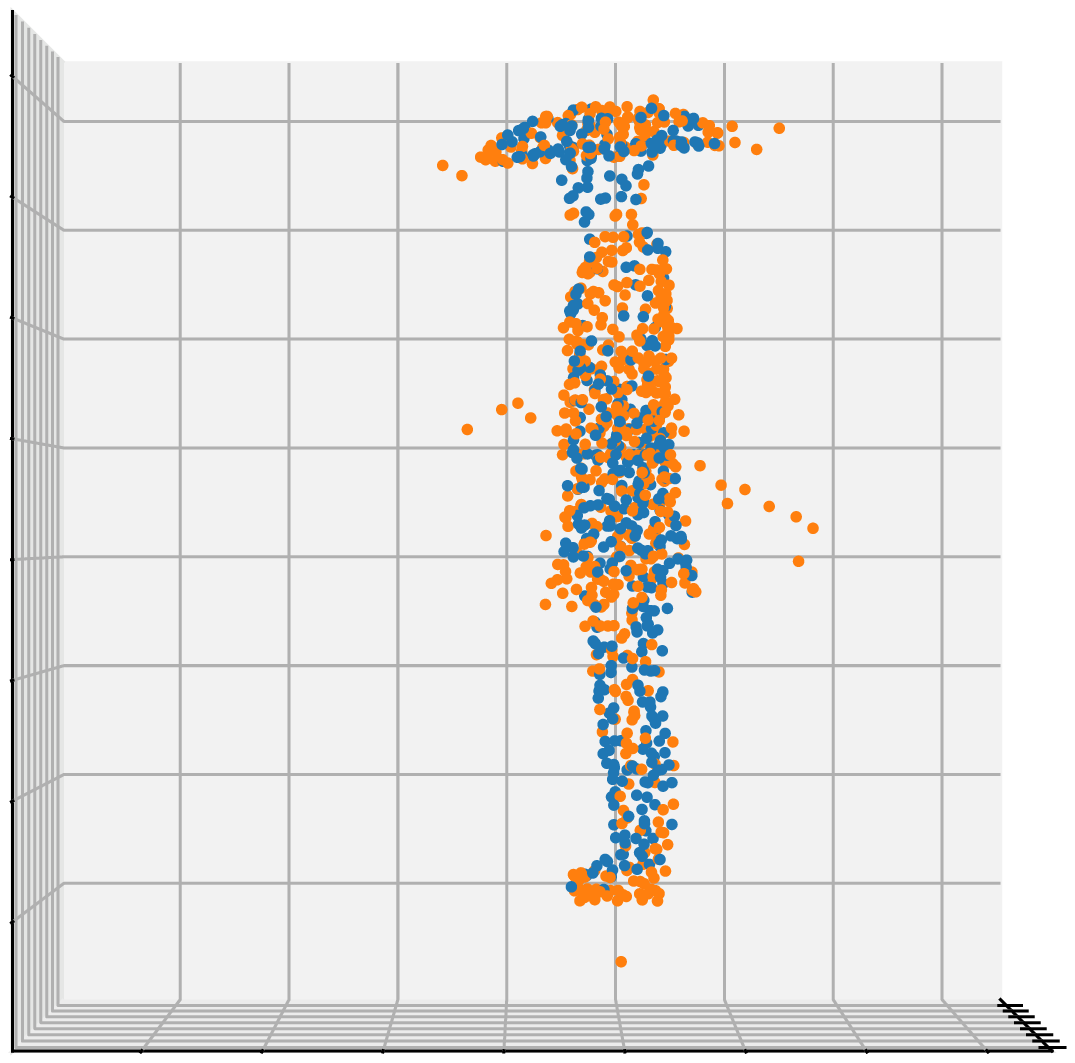}}
    \subfloat[Iter. gradient $L_2$ and clipping norms, predicted as plant.]{\includegraphics[trim={8cm 3cm 8cm 3cm}, clip=true, width=0.25\linewidth]{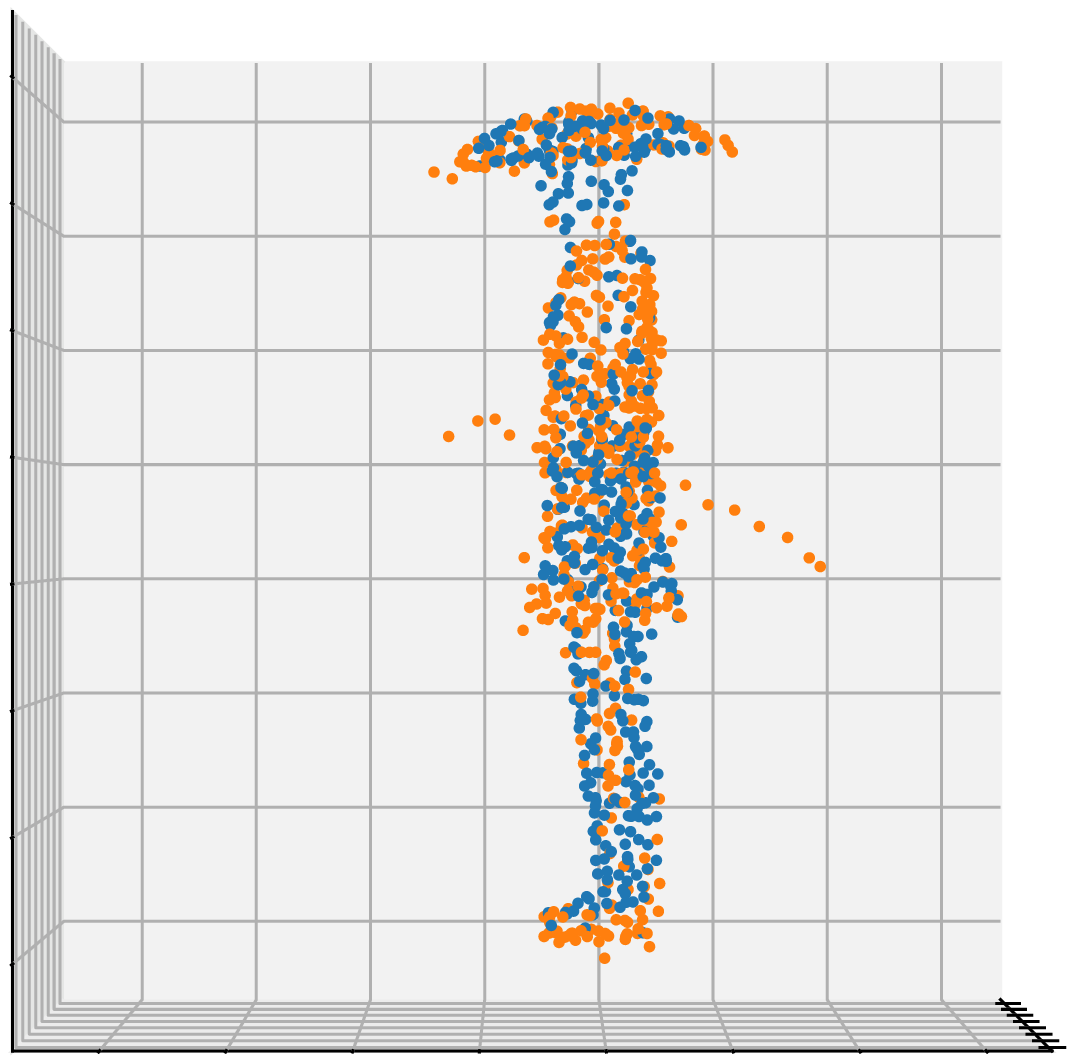}}
    \vfill
    \subfloat[Iter. gradient $L_2$ and gradient proj., predicted as plant.]{\includegraphics[trim={8cm 3cm 8cm 3cm}, clip=true, width=0.25\linewidth]{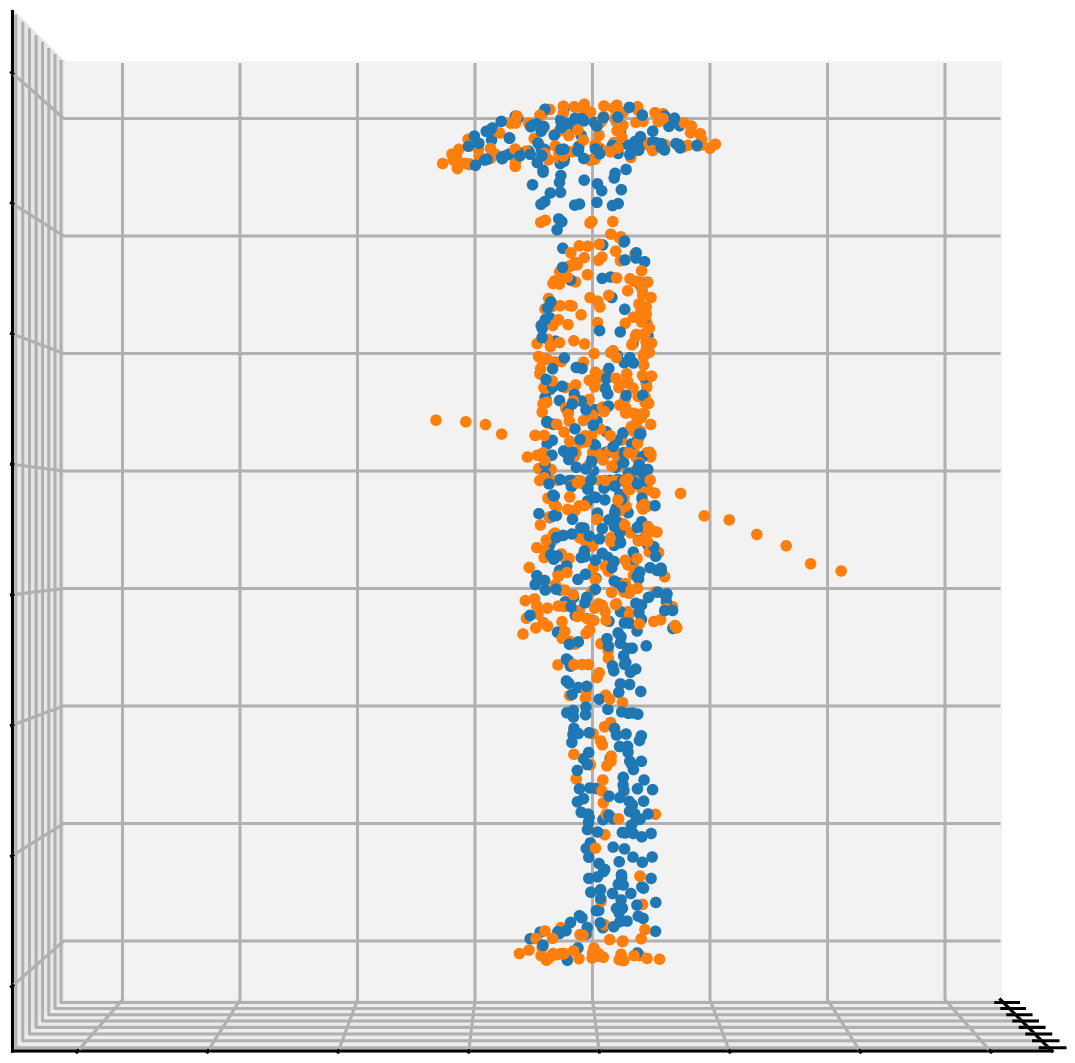}}
    \subfloat[Normalized fast gradient $L_2$, predicted as plant.]{\includegraphics[trim={8cm 3cm 8cm 3cm}, clip=true, width=0.25\linewidth]{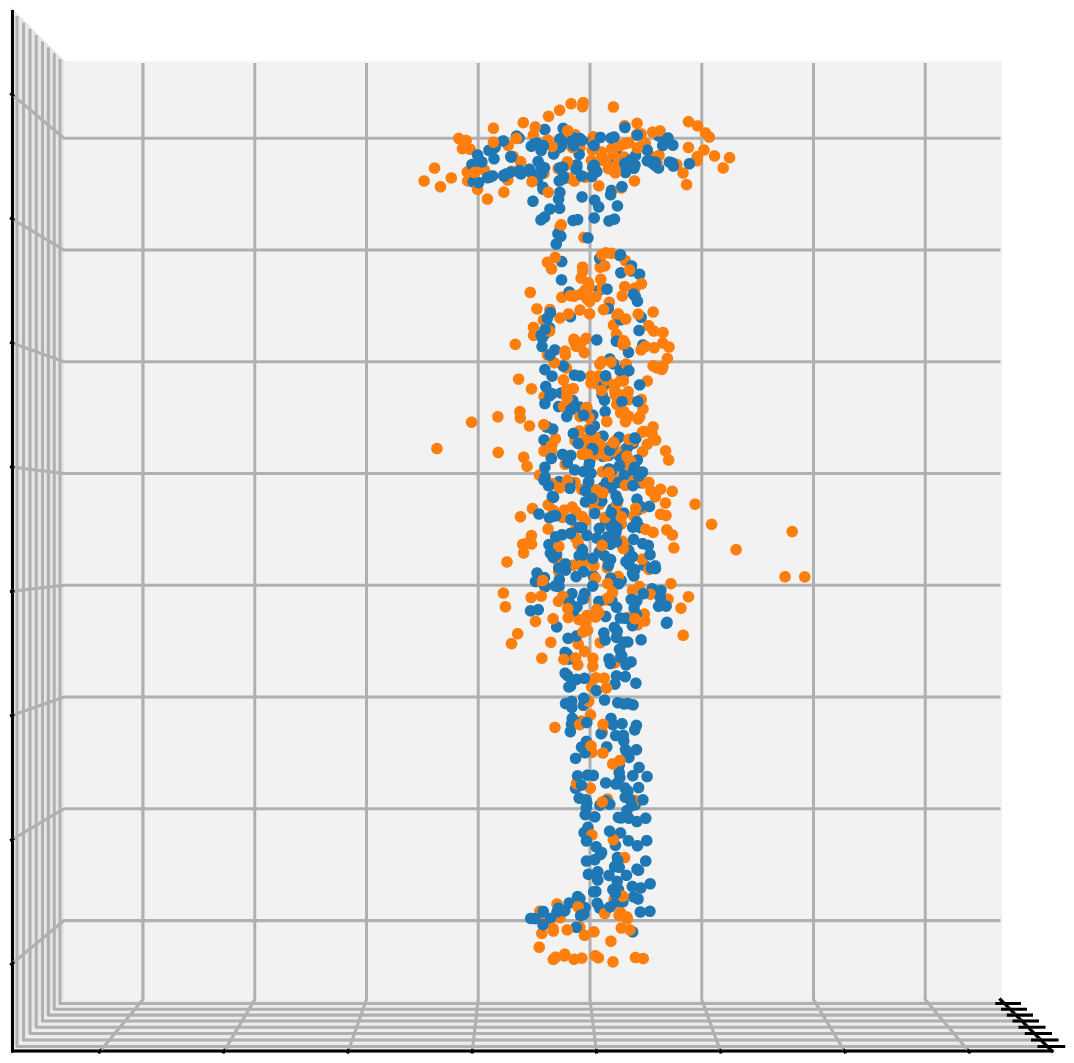}}
    \subfloat[Normalized iter. gradient $L_2$, predicted as plant.]{\includegraphics[trim={8cm 3cm 8cm 3cm}, clip=true, width=0.25\linewidth]{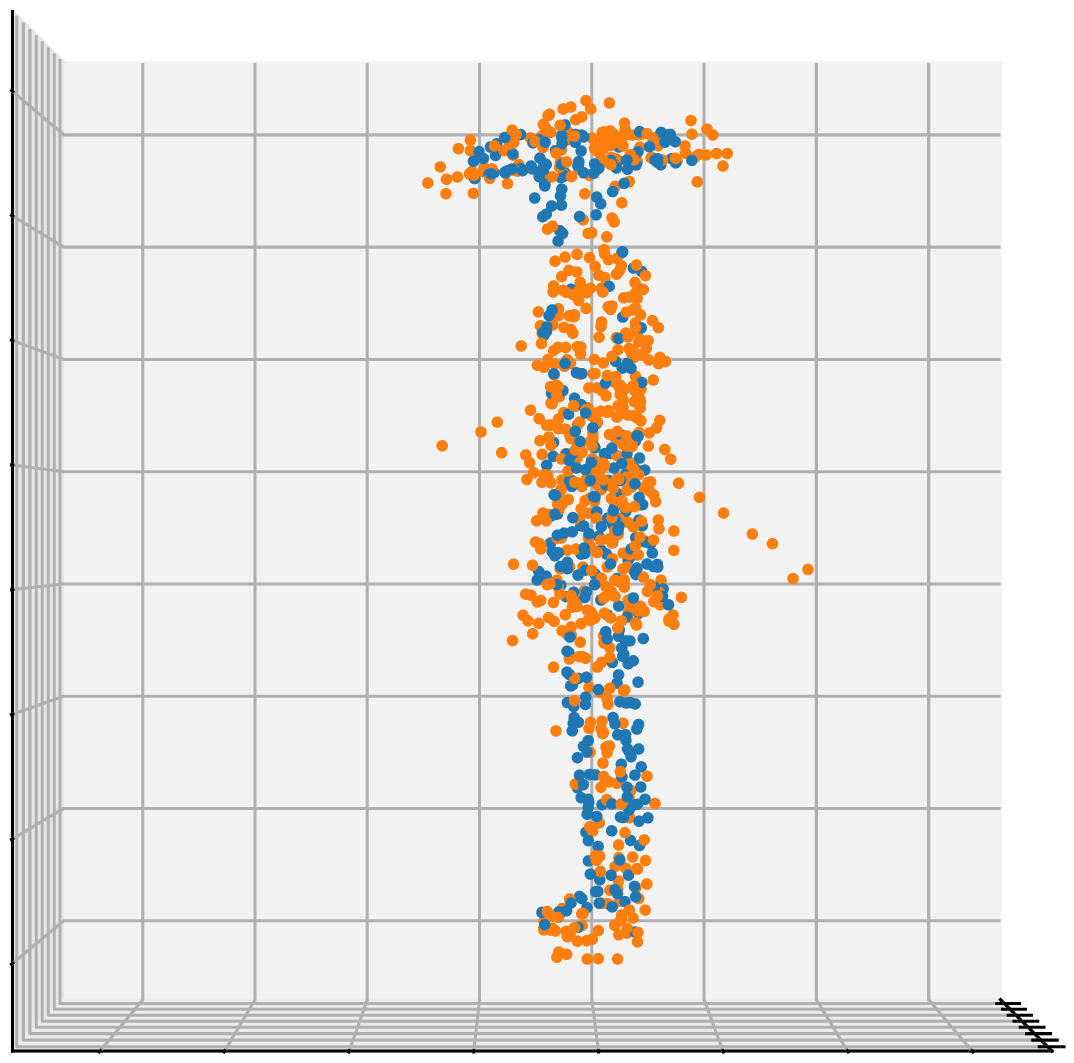}}
    \caption{A set of successful adversarial perturbations on the point cloud of a person, generated from PointNet trained with the ModelNet-Unique dataset. Orange points have nonzero perturbations.}
    \label{fig:adv_atk_viz_person}
\end{figure*}

\subsection{Jacobian-based saliency map (JSMA)}
As all of the aforementioned attacks act on all points within each point cloud, we also evaluate a different approach that selects a subset of points to perturb instead of perturbing all of the points. More specifically, we use a slightly modified version of the Jacobian-based saliency method~\cite{papernot2016limitations} for 3D points. The original paper describes a targeted attack that is constrained by the $L_0$ norm. \Ie, the attack seeks to minimize the number of input dimensions perturbed. We use an untargeted variant of the original attack.

Instead of selecting single dimensions to perturb, we aggregate the gradients for all three dimensions of each point and perturb them all. Furthermore, we allow dimensions to either increase or decrease in value depending on its gradients, which increases the number of candidate points that can be chosen to be perturbed. To craft the adversarial attack, we obtain the saliency $s(p)$ for each point $p$ through
\begin{align}
g_y(p) &= \frac{\partial f_y(x;\theta)}{\partial p}&
g_o(p) &= \sum_{i \neq y} \frac{\partial f_i(x;\theta)}{\partial p}
\end{align}
\begin{equation}
s(p) =
\begin{cases}
0 & \text{if } g_y(p), g_o(p) < 0\\
& \text{or } g_y(p), g_o(p) > 0\\
\sum (|g_y(p)| \odot |g_o(p)|) & \text{otherwise}
\end{cases}
\end{equation}
where the $\odot$ symbol denotes element-wise multiplication. We let the attack run for a fixed number of iterations, and for each iteration, the point with the highest total saliency across all dimensions is chosen. Then, each dimension of the chosen point's position is either increased or decreased by some $\epsilon$ according to whether $g_y(p) < g_o(p)$ or $g_y(p) > g_o(p)$, respectively.

\subsection{Other approaches}
One main problem with using adversarial attacks in 3D space is that, unlike 2D space, the perturbations are more perceptible due to obvious outliers that change the overall shapes of the point clouds. As such, in addition to those basic attacks, we also propose methods that reduce the perceptibility of those attacks.

\paragraph{Gradient projection.}
In this method, perturbations are projected onto the surface of an object, which is made up of a mesh of triangles. First, each adversarially perturbed point $p^{adv}$ is projected onto the plane represented by the triangle that point was sampled from:
\begin{equation}
p^{adv}_{proj} = p^{adv} - \vec{n}[\vec{n} \cdot (p^{adv} - t_1)]
\end{equation}
where $t_1$ is a vertex of the triangle, $\vec{n}$ is the unit normal vector of the plane, and $\cdot$ represents the dot product operation.

Then, each perturbed point that leaves its corresponding triangle's edges is clipped to the edges of that triangle.

This method shows that we can generate adversarial attacks of a point cloud by simply changing the sampling density. It generates adversarial examples that have the same general shape as the clean data, which makes the perturbations much less perceptible. It also allows us to measure how well the networks perform against changes in the distribution of points on an object's surface.

\paragraph{Clipping norms.}
A more practical way to lower the perceptibility of attacks (without requiring the triangular meshes of the point clouds) is to clip the $L_2$ norms of the perturbation of each point in order to match the average pairwise euclidean distances between nearby points in the clean sample. This limits large, outlying perturbations that may occur in one of the basic attacks.

To clip the $L_2$ norm, we use the following method for each point:
\begin{equation}
p_{clip}^{adv} = p + \frac{p^{adv} - p}{||p^{adv} - p||_2} \Big(\frac{1}{N} \sum_{p' \in x}||p' - \operatorname{NN}(p')||_2\Big)
\end{equation}
where $\operatorname{NN}(\cdot)$ returns the nearest neighboring point, and $N$ is the number of points in $x$. This attempts to maintain the distribution of distances between points in the perturbed sample to match that of the clean sample.
\begin{table*}[!htbp]
    \begin{center}
    \begin{tabular}{lcccc}
    \hline
    & None & Adversarial training & Removing outliers & Removing salient points\\
    \hline
    None & 0.0\% & 0.5\% & 2.6\% & 0.7\%\\
    Fast gradient $L_2$ & 39.8\% & 7.3\% & 4.6\% & 10.2\%\\
    Iter. gradient $L_2$ & 74.2\% & 37.1\% & 16.2\% & 19.9\%\\
    Iter. gradient $L_2$, clip norm & 45.2\% & 32.6\% & 10.9\% & 14.2\%\\
    Iter. gradient $L_2$, grad. proj. & 4.3\% & 6.8\% & 2.5\% & 2.1\%\\
    Normalized fast gradient $L_2$ & 14.1\% & 7.9\% & 9.1\% & 12.9\%\\
    Normalized iter. gradient $L_2$ & 64.5\% & 59.1\% & 19.7\% & 32.2\%\\
    JSMA & 40.9\% & 9.0\% & 0.5\% & 8.0\%\\
    \hline
    \end{tabular}
    \end{center}
    \caption{Error rates for untargeted attacks and defenses on PointNet trained with ModelNet-Unique. Each column represents a defense method, and each row represents an attack method.}
    \label{table:pointnet16_atk_def}
\end{table*}
\section{Defenses}
We evaluate the performance of several simple defensive techniques in response to the adversarial attacks. In addition to evaluating adversarial training, we also propose two different input restoration methods that try to remove perturbed points by making certain assumptions about clean input point clouds.

\subsection{Adversarial training}
The adversarial training algorithm was initially proposed by Goodfellow \etal~\cite{goodfellow2014explaining}. We train each model from scratch by generating fast gradient $L_2$ adversarial examples at each iteration and averaging the loss from feeding in batches of clean and adversarial examples:
\begin{equation}
\begin{aligned}
J^{adv}(f(x; \theta), y) &= \frac{1}{2}\Big[J\big(f(x; \theta), y\big)\ +\\
&J\big(f(x + \epsilon \frac{\Delta_x J(f(x; \theta), y)}{||\Delta_x J(f(x; \theta), y)||_2}; \theta), y\big)\Big]
\end{aligned}
\end{equation}
This trains the network to adjust for adversarial samples.

\subsection{Input restoration}
\paragraph{Removing outliers.}
Another way to defend against adversarial attacks is by removing outlying points that may be created due to adversarial perturbations. This is similar ideas used by~\cite{zhou2018deflecting}.

Outliers are identified by first examining the average euclidean distance of each point to its $k$-nearest neighbors:
\begin{equation}
d = \Big\{\frac{1}{k}\sum_{i = 1}^k ||p^{adv} - \operatorname{kNN}(p^{adv}, i)||_2\ \Big|\ p^{adv} \in x^{adv}\Big\}
\end{equation}
where $\operatorname{kNN}(\cdot, i)$ returns the $i$-th closest point.

Then, points that have very high average distances to its nearest neighbors are assumed to be outliers and are discarded. Each of these outlier points $p^{adv}$ are identified by looking at the distribution of average distances across all points:
\begin{equation}
\frac{1}{k}\sum_{i = 1}^k ||p^{adv} - \operatorname{kNN}(p^{adv}, i)||_2 > \frac{1}{|d|}\sum_i d_i + \epsilon \operatorname{stddev}(d)
\end{equation}
This method assumes that since each point on a natural shape should be uniformly sampled along the surface, any outlier point must be the result of adversarial perturbations.

\paragraph{Removing salient points.}
We also explore a defensive technique supported by the crude assumption that perturbed points should have relatively large magnitudes of gradients. By assuming that this is true, an algorithm that discards the most salient points can be used, where the saliency of each point $p^{adv}$ is given by
\begin{equation}
s(p^{adv}) = \max_i \bigg|\bigg|\frac{\partial f_i(x^{adv};\theta)}{\partial p^{adv}}\bigg|\bigg|_2
\end{equation}
This method may remove both unperturbed points and perturbed points.
\section{Evaluation}
\subsection{Models}
We evaluate both PointNet \cite{qi2017pointnet} and PointNet++ \cite{qi2017pointnetplusplus} for their performance against the mentioned adversarial attacks and defenses. We directly use the default hyperparameters when training the networks, except for a slightly lower batch size for PointNet++ due to limited memory.

\subsection{Datasets}
We use shapes from the ModelNet-40 \cite{wu2015shapenet} dataset to train and evaluate the models. There are over 2400 total models from 40 different classes in the dataset.
We sample 1024 points from each shape and center and scale the data to match the settings used by Qi \etal~\cite{qi2017pointnet, qi2017pointnetplusplus} in PointNet and PointNet++.

Since some of the classes in ModelNet-40 are quite indistinguishable even to humans (\eg chair and stool), for most experiments, we use a subset of 16 hand-picked object classes that have more unique shapes, which 
allows us to measure the effectiveness of adversarial attacks that have to switch between very different classes. The 16 classes are: airplane, bed, bookshelf, car, chair, cone, cup, guitar, lamp, laptop, person, piano, plant, range hood, stairs, and table. We will refer to this dataset as ModelNet-Unique.

\subsection{Implementation details}
For all attacks that constrain the $L_2$ norm between the clean and adversarial point clouds, we use an $\epsilon$ value of 1. For normalized fast/iterative gradient attacks, we use an $\epsilon$ value of 0.05. Finally, for JSMA, we use 0.5 as the $\epsilon$ value. We also use 10 iterations for all iterative attacks, including JSMA. 

For our targeted iterative gradient $L_2$ attacks, we use a higher $\epsilon$ of 5, as the difficulty of generating successful adversarial attacks is increased.

For our evaluations of the defensive techniques, we adversarially train with perturbations generated by fast gradient $L_2$ using an $\epsilon$ value of 1.
For the outlier removal method, we use the mean distance to the 10 closest neighbors of each point and we clip perturbations that exceed the mean by 1 standard deviation. We remove 100 of the most salient points when removing salient points.

\section{Results}
\subsection{Clean inputs}
We perform all of our attacks on only the correctly classified point clouds. For PointNet and PointNet++ using the full 40 classes, around 90\% of the point clouds are correctly classified. On ModelNet-Unique, around 96\%  of the point clouds are correctly classified.

\subsection{Effectiveness of white-box attacks and defenses}
\begin{table}[!htbp]
\begin{center}
\begin{tabular}{lc}
\hline
& Success rate\\
\hline
Fast gradient $L_2$ & 58.8\%\\
Iter. gradient $L_2$ & 90.1\%\\
Iter. gradient $L_2$, clip norm & 77.0\%\\
Iter. gradient $L_2$, gradient proj. & 26.0\%\\
Normalized fast gradient $L_2$ & 40.0\%\\
Normalized iter. gradient $L_2$ & 88.1\%\\
JSMA & 56.4\%\\
\hline
\end{tabular}
\end{center}
\caption{Success rates for untargeted attacks on PointNet trained with ModelNet-40.}
\label{table:pointnet40_atk}
\end{table}
\begin{table}[!htbp]
    \begin{center}
    \begin{tabular}{lcc}
    \hline
    & 40 & Unique\\
    \hline
    Fast gradient $L_2$ & 36.5\% & 36.1\%\\
    Iter. gradient $L_2$ & 96.4\% & 92.2\%\\
    Iter. gradient $L_2$, clip norm & 91.2\% & 70.6\%\\
    Iter. gradient $L_2$, grad. proj. & 24.5\% & 4.6\%\\
    Normalized fast gradient $L_2$ & 31.0\% & 24.7\%\\
    Normalized iter. gradient $L_2$ & 96.6\% & 91.6\%\\
    JSMA & 9.8\% & 2.5\%\\
    \hline
    \end{tabular}
    \end{center}
    \caption{Success rates of untargeted attacks on PointNet++. The network is trained/evaluated on both ModelNet-40 (40) and ModelNet-Unique (Unique).}
    \label{table:pointnet2_atk}
\end{table}
For ModelNet-Unique, the error rates for all combinations of attacks and defenses on PointNet are shown in Table \ref{table:pointnet16_atk_def}. Results for ModelNet-40 on PointNet are shown in Table \ref{table:pointnet40_atk}. Table \ref{table:pointnet2_atk} shows the results on PointNet++, for both ModelNet-Unique and ModelNet-40.

We also show a few visualizations of adversarial examples generated from PointNet on the ModelNet-Unique set in Figure \ref{fig:adv_atk_viz_car} and Figure \ref{fig:adv_atk_viz_person}.

Adversarial attacks on undefended networks are extremely effective against PointNet and PointNet++. Furthermore, PointNet++ has higher error rates than PointNet for both the vanilla and the normalized versions of the iterative gradient $L_2$ attack, even though it is more complex, which suggests that higher architecture complexity does not lead to higher robustness against adversarial attacks. However, PointNet++ shows greater resistance to JSMA, which we think is due to how it hierarchically groups relatively close features within epsilon balls, allowing it to ignore large perturbations.

The defenses we evaluate are also effective. Adversarial training halves the success rate of iterative gradient $L_2$, and decreases the success rates of fast gradient $L_2$ and JSMA by more than 4 times. However, it is much less effective against normalized iterative gradient $L_2$ compared to iterative gradient $L_2$. This suggests that adversarial training does not transfer very well to perturbations that have different distributions.

Overall, the other two simpler defenses perform even better than adversarial training. We find that both removing outliers and removing salient points, which were constructed to defend against large perturbations, are also effective against attacks, like $L_2$ norm clipping and gradient projection, that generate small perturbations. Furthermore, directly removing salient points does not damage the classification of clean input point clouds by too much compared to other methods.

The best defensive method is by removing outliers, at the expense of 2.6\% lower accuracy on unperturbed inputs.

As expected, adversarial attacks are more successful on ModelNet-40 than on ModelNet-Unique.

\paragraph{Perceptibility.}
\begin{table}[!htbp]
\begin{center}
\begin{tabular}{lcc}
\hline
& $||x^{adv} - x||_2$\\
\hline
Fast gradient $L_2$ & 1.0\\
Iter. gradient $L_2$ & 0.6\\
Iter. grad. $L_2$, clip norm & 0.4\\
Iter. grad. $L_2$, grad. proj. & 0.2\\
Normalized fast. grad $L_2$ & 1.2\\
Normalized iter. grad $L_2$ & 0.7\\
JSMA & 2.6\\
\hline
\end{tabular}
\end{center}
\caption{Average $L_2$ norms of the adversarial perturbations ($||x^{adv} - x||_2$) for PointNet trained with ModelNet-Unique.}
\label{table:pointnet16_l2_norm}
\end{table}
The iterative gradient $L_2$ attack with gradient projection is the least perceptible attack. However, It reaches over 20\% success rate on both PointNet and PointNet++ with the ModelNet-40 dataset, even though there are barely any visible change to the input point clouds. Furthermore, the predictions of successful adversarial attacks are all highly confident. With higher epsilons and slightly more noticeable perturbations, the success rate of the attack plateaus at around 30\% to 40\% for both PointNet and PointNet++ on the ModelNet-40. The second least perceptible attack, iterative gradient $L_2$ with clipping norms, is much more successful as it reaches a 45.2\% attack success rate on PointNet with the harder ModelNet-Unique dataset, and even higher on PointNet++. This clearly indicates the vulnerability of those networks against adversarial attacks that are almost imperceptible.

We show the $L_2$ norms between the clean point clouds and the adversarial point clouds in Table~\ref{table:pointnet16_l2_norm}. This can be seen as a way of measuring perceptibility, though it does not account for the shapes of the point clouds.

\subsection{Targeted white-box attacks}
\begin{figure}[!htbp]
    \includegraphics[trim={0 0 3.5cm 2.5cm}, clip=true,width=0.47\textwidth]{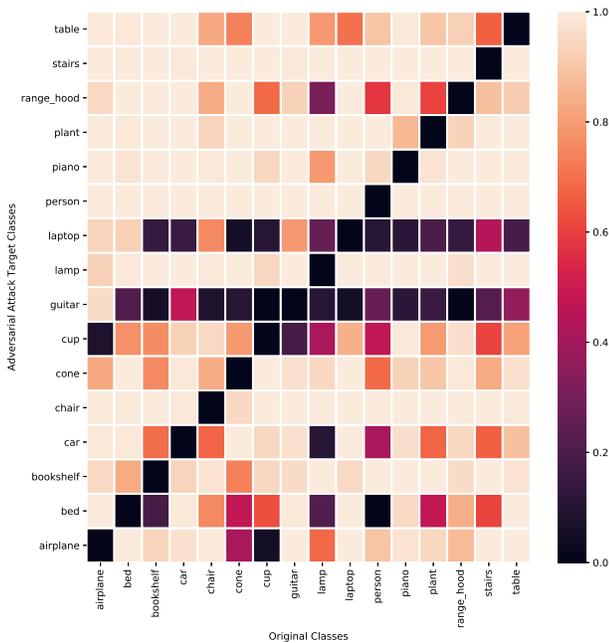}
    \caption{A heat map of successful targeted iterative gradient $L_2$ attacks on PointNet for every pair of classes from the ModelNet-Unique dataset. The x-axis represents the labels and the y-axis represents attack target. Brighter is more successful. Each cell represents the percentage of successful attacks.}
    \label{fig:targeted_heatmap}
\end{figure}
We show a heat map in Figure \ref{fig:targeted_heatmap} that describes how well targeted iterative gradient $L_2$ attacks perform when each clean point cloud targets every output class. Overall, the targeted attacks are very successful, even though the difficulty level for crafting adversarial examples is increased. The average success rate for all targeted attacks is 79.6\%. The average prediction confidence of 97.0\% for successful attacks.

\subsection{Effectiveness of black-box attacks}
\begin{table}[!htbp]
    \begin{center}
    \begin{tabular}{lcc}
    \hline
    & Fast grad. $L_2$ & Iter. grad. $L_2$\\
    \hline
    PN $\rightarrow$ PN++ & 30.6\% & 14.7\%\\
    PN++ $\rightarrow$ PN & 29.3\% & 10.0\%\\
    PN adv. train $\rightarrow$ PN & 62.0\% & 36.0\%\\
    PN $\rightarrow$ PN adv. train & 11.8\% & 11.7\%\\
    \hline
    \end{tabular}
    \end{center}
    \caption{Success rates of transfer attacks between regular and adversarially trained PointNet (PN) and PointNet++ (PN++), on the ModelNet-Unique dataset. Only successful perturbations from one network were evaluated on the other network. \Ie, each percentage represents a fraction of all successful attacks from one model that were successful in fooling another model.}
    \label{table:transfer_atk}
\end{table}
We evaluate the effectiveness of black-box transfer attacks by generating adversarial perturbations for each network architecture and testing them on the other network. Our results are shown in Table \ref{table:transfer_atk}.

Our results are consistent with the results for black-box attacks on 2D images \cite{kurakin2016adversarialscale}, as fast gradient $L_2$ performs better than iterative gradient $L_2$. This was determined by \cite{kurakin2016adversarialscale} to be due to iterative gradient $L_2$ attacks overfitting the model it was crafted for.

Adversarially trained PointNet shows resistance to adversarial perturbations generated from a normally trained PointNet architecture. Also, attacks that can fool an adversarially trained PointNet have a larger success rate on fooling an undefended PointNet\footnote{Note that success rate of fast gradient $L_2$ transferring from PointNet with adversarial training to the undefended PointNet is slightly statistically insignificant, as less than 100 adversarial examples successfully fooled an adversarially trained PointNet}. These results are expected, and they show that adversarial training is robust to simple transfer attacks from the same model architecture.
\section{Discussion}
PointNet and PointNet++ have been shown to be robust against point clouds of varying densities and randomly perturbed point clouds~\cite{qi2017pointnet, qi2017pointnetplusplus}. However, against our adversarial attacks that preserve the overall shape of the input point clouds, the networks perform very poorly. This indicates a fundamental problem with the decision boundaries that PointNet and PointNet++ learn. The latent representations of input point clouds must be close to the decision boundaries if they can easily cross it with small perturbations.

Our defenses attempts to lessen the problem. Adversarial training forces the decision boundaries to adjust for adversarial perturbations, leaving adequate space between the clean sample and the decision boundary~\cite{madry2017towards}. Removing outliers does not directly affect a network's learned parameters, but it moves the input data away from the decision boundaries by enforcing certain distributions for the input data. \Ie, it restores the point clouds to a state where each point is relatively close to its neighbors, which should be true for clean point clouds that were uniformly sampled from 3D meshes. Removing salient points attempts to drop points that may significantly affect the network's prediction, which also moves the point clouds away from the decision boundaries.

The max-pooling operations in PointNet and PointNet++ hide a subset of points from attacks that require gradient information. The attacks cannot perturb those hidden points because they have zero gradients due to not being selected by the max-pooling operation. As outliers and salient points are removed, points that were previously hidden by the max-pooling operation are exposed and they can represent the overall shape of the point clouds, allowing the networks to make accurate predictions.

The existence of certain priors in the distribution of unperturbed points, the hiding of points by the max-pooling operation, and the ability to directly drop points that may be perturbed without disturbing the overall shapes of the point clouds allow 3D point clouds and 3D point cloud networks to be inherently more robust against adversarial attacks than 2D images and 2D convolutional networks.

We think that our outlier removal method provides a necessary upper bound for future evaluations of adversarial attacks in 3D space, as unlike image pixels, each point can be perturbed by an arbitrary amount. Removing very obvious outliers is necessary to prevent attacks on 3D point clouds that may create effective, but unrealistic changes to the input data.
\section{Conclusion}
We conduct a preliminary examination on adversarial attacks and defenses on 3D point clouds point cloud classifiers, like PointNet~\cite{qi2017pointnet} and PointNet++~\cite{qi2017pointnetplusplus}. We show that many methods used on 2D images are also effective on 3D point clouds. In addition, we propose various methods to reduce the perceptibility of adversarial perturbations while remaining relatively effective, and we examine simple defenses that exploit the nature of the 3D point cloud data.

Overall, we find that although deep 3D point cloud classifiers are still susceptible to simple gradient-based adversarial attacks, they are more easily defensible compared to 2D image classifiers.

As deep neural networks are applied to various problems, the significance of adversarial examples grows. We hope that our work can provide a foundation for further research into understanding network behavior in an adversarial setting, and improving the robustness of neural networks that handle 3D data in safety-critical applications.

{\small
\bibliographystyle{ieee}
\bibliography{main}
}

\end{document}